\documentclass[10pt,twocolumn,letterpaper]{article}

\usepackage{wacv}
\usepackage{times}
\usepackage{epsfig}
\usepackage{graphicx}
\usepackage{amsmath}
\usepackage{amssymb}

\usepackage{threeparttable}  
\usepackage{paralist}
\usepackage{verbatim}
\usepackage{comment}
\usepackage{amsmath}
\usepackage{amssymb}
\usepackage{subfig}
\usepackage[toc,page]{appendix}

\wacvfinalcopy 

\def\wacvPaperID{178} 

\ifwacvfinal\fi
\begin{document}

\title{Satellite Imagery Multiscale Rapid Detection with Windowed Networks}

\author{Adam Van Etten \\
In-Q-Tel CosmiQ Works\\
{\tt\small avanetten@iqt.org}
}

\maketitle
\ifwacvfinal\thispagestyle{empty}\fi

\begin{abstract}
Detecting small objects over large areas remains a significant challenge in satellite imagery analytics.  
Among the challenges is the sheer number of pixels and geographical extent per image: a single DigitalGlobe satellite image encompasses over 64 km$^2$ and over 250 million pixels. Another challenge is that objects of interest are often minuscule ($\sim10$ pixels in extent even for the highest resolution imagery), which complicates traditional computer vision techniques. 
To address these issues, we propose a pipeline (SIMRDWN) that evaluates satellite images of arbitrarily large size at native resolution
at a rate of $\geq 0.2$ $\rm{km}^2 / \rm{s}$.
Building upon the 
tensorflow object detection API paper \cite{tf_obj_det},
this pipeline offers a unified approach to multiple object detection frameworks that can run inference on images of arbitrary size.  The SIMRDWN pipeline includes a modified version of YOLO (known as YOLT \cite{yolt}), along with the models in \cite{tf_obj_det}: SSD \cite{ssd}, Faster R-CNN \cite{faster_rcnn}, and R-FCN \cite{rfcn}.
The proposed approach allows comparison of the performance of these four frameworks, and can rapidly detect objects of vastly different scales with relatively little training data over multiple sensors.  For objects of very
different scales (e.g.\ airplanes versus airports) we find that using two different detectors at different scales is very effective with negligible runtime cost.
We  evaluate large test images at native resolution and find mAP scores of 0.2 to 0.8 for vehicle localization, with the YOLT architecture achieving both the highest mAP and fastest inference speed.

\end{abstract}

\section{Introduction}\label{sec_intro}

Computer vision techniques have made great strides in the past few years since the introduction of convolutional neural networks \cite{alexnet} in the ImageNet \cite{imagenet} competition. The availability of large, high-quality labeled datasets such as ImageNet \cite{imagenet}, PASCAL VOC \cite{pascal_voc} and MS COCO \cite{ms_coco} have helped spur a number of impressive advances in rapid object detection that run in near real-time;  four of the best are: Faster R-CNN \cite{faster_rcnn}, R-FCN \cite{rfcn}, SSD \cite{ssd}, and YOLO \cite{yolov1},\cite{yolo9000}. 
Faster R-CNN and R-FCN
typically ingests $1000\times600$ pixel images \cite{faster_rcnn}, \cite{rfcn}, whereas SSD uses $300\times300$ or $512\times512$ pixel input images \cite{ssd}, and YOLO runs on either $416\times416$ or $544\times544$ pixel inputs \cite{yolo9000}.  While the performance of all these frameworks is impressive, none can come remotely close to ingesting the $\sim16,000\times16,000$ input sizes typical of satellite imagery.  
The speed and accuracy tradeoffs of Faster-RCNN, R-FCN, and SSD were compared in depth in \cite{tf_obj_det}.  Missing from these comparisons was the YOLO framework, which has demonstrated competitive scores on the PASCAL VOC dataset, along with high inference speeds.  
 The YOLO authors also showed that this framework is highly transferrable to new domains by demonstrating 
superior performance to other frameworks (i.e.\ SSD and Faster R-CNN) on the Picasso Dataset \cite{picasso_dataset} and the People-Art Dataset \cite{people-art_dataset}.  In addition an extension of the YOLO framework (dubbed YOLT for You Only Look Twice) \cite{yolt} showed promise in localizing objects in satellite imagery.
The speed, accuracy, and flexibility of YOLO therefore merits a full comparison with the other three frameworks, and motivates this study.  


The application of deep learning methods to traditional object detection pipelines is non-trivial for a variety of reasons. The unique aspects of satellite imagery necessitate algorithmic contributions to address challenges related to the spatial extent of foreground target objects, complete rotation invariance, and a large scale search space. Excluding implementation details, algorithms must adjust for:

\begin{description}
\item[Small spatial extent] In satellite imagery objects of interest are often very small and densely clustered, rather than the large and prominent subjects typical in ImageNet data. In the satellite domain, resolution is typically defined as the ground sample distance (GSD), which describes the physical size of one image pixel.  Commercially available imagery varies from 30 cm GSD for the sharpest DigitalGlobe imagery, to $3-4$ meter GSD for Planet imagery. This means that for small objects such as cars each object will be only $\sim15$ pixels in extent even at the highest resolution.
\item[Complete rotation invariance] Objects viewed from overhead can have any orientation (e.g.\ ships can have any heading between 0 and 360 degrees, whereas trees in ImageNet data are reliably vertical)
\item[Training example frequency] There is a relative dearth of training data (though efforts such as SpaceNet\footnote{https://aws.amazon.com/public-datasets/spacenet/} are attempting to ameliorate this issue)
\item[Ultra high resolution] Input images are enormous (often hundreds of megapixels), so simply downsampling to the input size required by most algorithms (usually a few hundred pixels) is not an option 
\end{description}

On the plus side,  one can leverage the relatively constant distance from sensor to object, which is well known and is typically $\sim400$ km. This coupled with the nadir facing sensor results in consistent pixel to metric ratio of objects.




Section \ref{sec_related_work} details in further depth the challenges faced by standard algorithms when applied to satellite imagery. The remainder of this work is broken up to describe the proposed contributions as follows. Section \ref{sec_algo} describes model architectures. With regard to rotation invariance and small labeled training dataset sizes, Section \ref{sec_train} describes data augmentation and size requirements.  Section \ref{sec_test} details the test dataset.  
Section \ref{sec_test_proc}  details the experiment design and our method for splitting, evaluating, and recombining large test images of arbitrary size at native resolution. Finally, the performance of the various algorithms is discussed in detail in Section~\ref{sec_results}.

\section{Related Work}\label{sec_related_work}


Many recent papers that apply advanced machine learning techniques to aerial or satellite imagery focus on a slightly different problem than the one we attempt to address.  For example, \cite{rs_obj_det_slow} showed good performance on localizing objects in overhead imagery; yet with an inference speed of $10 - 40$ seconds per $1280\times1280$ pixel image chip this approach will not scale to large area inference.  Efforts to localize surface to-air-missile sites \cite{samfinder} with satellite imagery and sliding window classifiers work if one only is interested in a single object size of hundreds of meters.  Running a sliding window classifier across the image to search for small objects of interest quickly becomes computationally intractable, however, since multiple window sizes will be required for each object size.  For perspective, one must evaluate over one million sliding window cutouts if the target is a 10 meter boat in a DigitalGlobe image. 

Efforts such as \cite{mnihroads}, \cite{roadunet} have shown success in extracting roads from overhead imagery via segmentation techniques.  Similarly, \cite{vakabuildings} extracted rough building footprints via pixel segmentation combined with post-processing techniques; such segmentation approaches are quite different from the rapid object detection approach we propose.  

Application of rapid object detection algorithms to the remote sensing sphere is still relatively nascent, as evidenced by the lack of reference to SSD, Faster-RCNN, or YOLO in a recent survey of object detection in remote sensing \cite{rs_obj_det_survey}.  While tiling a large image is still necessary, the larger field of view of these frameworks (a few hundred pixels) compared to simple classifiers (as low as 10 pixels) results in a reduction in the number of tiles required by a factor of over 1000. This reduced number of tiles yields a corresponding marked increase in inference speed. In addition, object detection frameworks often have much improved background differentiation since the network encodes contextual information for each object.  

The rapid object detection frameworks of  YOLO, SDD, Faster-RCNN, R-FCN have significant runtime advantages to other methods detailed above, yet complications remain.  For example, small objects in groups, such as flocks of birds present a challenge ~\cite{yolov1}, caused in part by the multiple downsampling layers of the convolutional networks.  Further, these multiple downsampling layers result in relatively coarse features  for object differentiation; this poses a problem if objects of interest are only a few pixels in extent.  For example, consider the default YOLO network architecture, which downsamples by a factor of 32 and returns a $13\times13$ prediction grid \cite{yolo9000}; this means that object differentiation is problematic if object centroids are separated by less than 32 pixels.  
Faster-RCNN downsamples by a factor of 16 by default \cite{faster_rcnn}, which in theory permits a higher density of object than the standard YOLO architecture.  SSD incorporates features at multiple downsampling layers to improve performance on small objects \cite{ssd}. R-FCN proposes 300 regions of interest, and then refines positions within that ROI via a $k \times k$ grid, where by default $k=3$.
Another difficulty for object detection algorithms applied to satellite imagery is that algorithms often struggle to generalize objects in new or unusual aspect ratios or configurations~\cite{yolov1}.  Since objects can have arbitrary heading, this limited range of invariance to rotation is troublesome.  

Our response is to leverage rapid object detection algorithms to evaluate satellite imagery with a combination of local image interpolation and a multiscale ensemble of detectors.  Along with attempting to address the issues listed above and in Section \ref{sec_intro}, we spend significant effort comparing how well SSD, Faster-RCNN, RFCN, and YOLO/YOLT perform when applied to satellite imagery.

\section{SIMRDWN}\label{sec_algo}


In order to address the limitations discussed in Section \ref{sec_related_work}, we implement 
an object detection framework optimized for overhead imagery: 
Satellite Imagery Multiscale Rapid Detection with Windowed Networks (SIMRDWN).
We extend the Darknet neural network framework \cite{darknet} and update a number of the C libraries to enable analysis of geospatial imagery and integrate with external python libraries \cite{yolt}.  We combine this modified Darknet code with the Tensorflow object detection API \cite{tf_obj_det} to create a unified framework.  Current rapid object detection frameworks can only infer on images a few hundred pixels in size; since our framework is designed for overhead imagery we implement techniques to analyze test images of arbitrary size.  

\subsection{Large Image Inference}\label{sec_infer}
\begin{figure}
\begin{center}
\includegraphics[height=7cm]{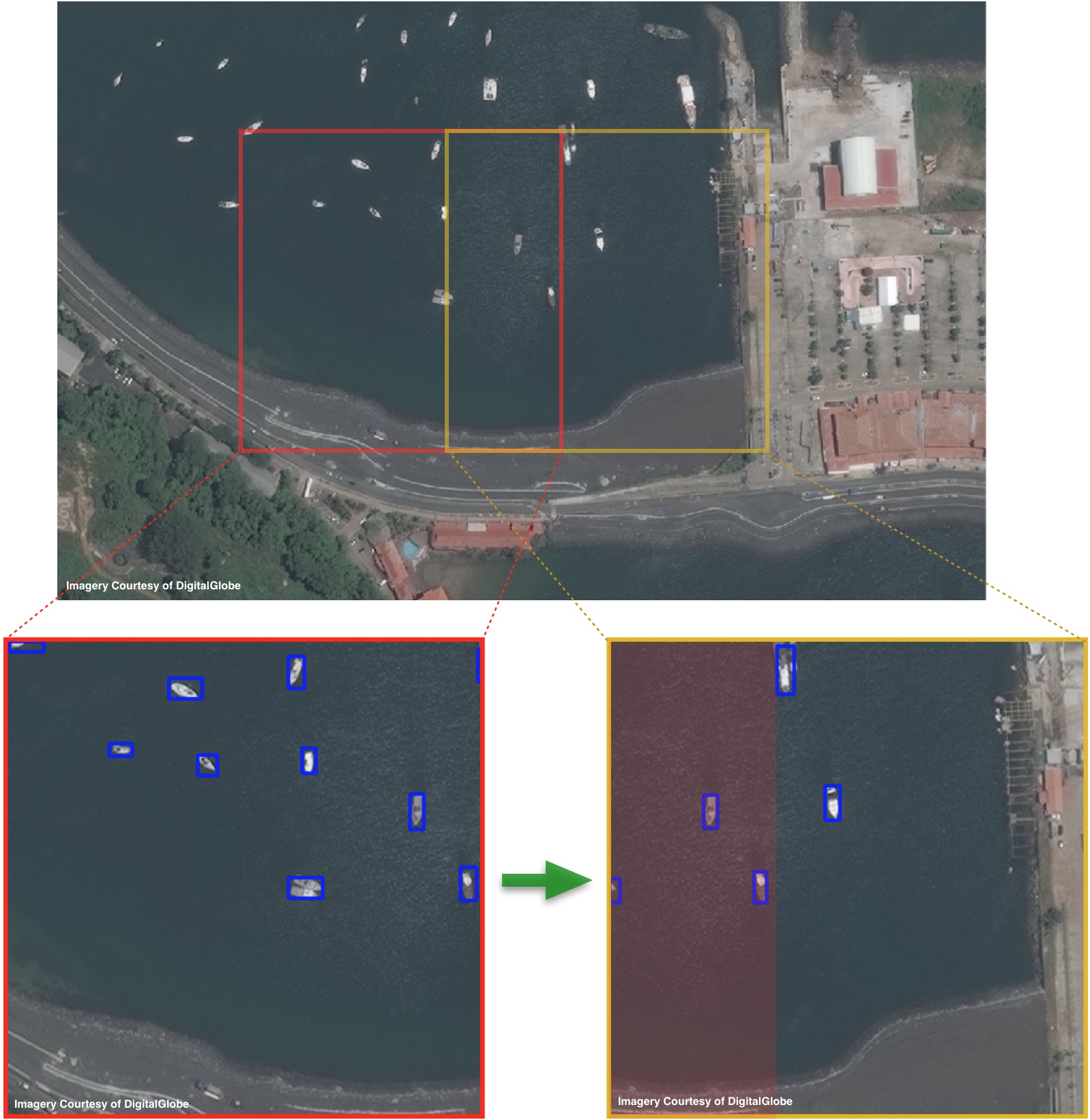}
\end{center}
\caption{Example of 416 pixel sliding window going from left to right across a large test image.
The overlap of the bottom right image is shown in red.  Non-maximal suppression of this overlap is necessary to refine detections at the edge of the cutouts where objects may be truncated by the window boundary.}
\label{fig:slid_win}
\end{figure}

We 
partition testing images of arbitrary size into manageable cutouts (416 pixels be default) and run each cutout through the trained models. We refer to this process as windowed networks.
Partitioning takes place via a sliding window with user defined bin sizes and overlap ($15\%$ by default),
see Figure \ref{fig:slid_win}.  We record the position of each sliding window cutout by naming each cutout according to the schema:

\texttt{ImageName|row\underline{{ }}column\underline{{ }}height\underline{{ }}width.ext}

\noindent
For example:

\texttt{panama50cm|1370\underline{{ }}1180\underline{{ }}416\underline{{ }}416.tif}

\subsection{Post-Processing}\label{sec_post_proc}

Much of the utility of satellite (or aerial) imagery lies in its inherent ability to map large areas of the globe.  Thus, small image chips are far less useful than the large field of view images produced by satellite platforms.  The final step in the object detection pipeline therefore seeks to stitch together the hundreds or thousands of testing chips into one final image strip.  

For each cutout the bounding box position predictions returned from the classifier are adjusted according to the \texttt{row} and \texttt{column} values of that cutout; this provides the global position of each bounding box prediction in the original input image.  The $15\%$ overlap ensures all regions will be analyzed, but also results in overlapping detections on the cutout boundaries.  We apply non-maximal suppression to the global matrix of bounding box predictions to alleviate such overlapping detections.

\subsection{Model Architectures}\label{sec_arch}

\begin{description}

\item[YOLO]
We follow the implementation of \cite{yolt} and utilize a modified Darknet \cite{darknet} framework to apply the standard YOLO configuration.    
We use the standard model architecture of YOLOv2 \cite{yolo9000}, which outputs a $13\times13$ grid.  Each convolutional layer is batch normalized with a leaky rectified linear activation, save the final layer that utilizes a linear activation. The final layer provides predictions of bounding boxes and classes, and has size:
$N_f = N_{\rm{boxes}} \times (N_{\rm{classes}} + 5)$, 
where $N_{\rm{boxes}}$ is the number of boxes per grid (5 by default), and $N_{\rm{classes}}$ is the number of object classes \cite{yolov1}.  
We train with stochastic gradient descent and maintain many of the hyper parameters of \cite{yolo9000}: 5 boxes per grid, an initial learning rate of $10^{-3}$, a weight decay of 0.0005, and a momentum of 0.9.  We use a batch size of 16 and train for 60,000 iterations.

\item[YOLT] 
To reduce model coarseness and accurately detect dense objects (such as cars), we follow \cite{yolt} and implement a network architecture that uses 22 layers and downsamples by a factor of 16 rather than the standard $32\times$ downsampling of YOLO. 
Thus, a $416\times416$ pixel input image yields a $26\times26$ prediction grid. 
Our architecture is inspired by the 28-layer YOLO network, though this new architecture is optimized for small, densely packed objects.  
The dense grid is unnecessary for diffuse objects such as airports, but improves performance for high density scenes such as parking lots; the fewer number of layers increases run speed.
To improve the fidelity of small objects, we also include a passthrough layer (described in \cite{yolo9000}, and similar to identity mappings in ResNet \cite{resnet}) that concatenates the final $52\times52$ layer onto the last convolutional layer, allowing the detector access to finer grained features of this expanded feature map.  We utilize the same hyperparameters as the YOLO implementation.  

\item[SSD] We follow the SSD implementation of \cite{tf_obj_det}.  We experiment with both Inception V2 \cite{inceptionv2} and MobileNet \cite{mobilenet} architectures.  
For both models we adopt a base learning rate of 0.004 and a decay rate of 0.95.
We train for 30,000 iterations with a batch size of 16, and use the 
``high-resolution'' setting of $600 \times 600$ pixel image sizes.  
These two SSD model architectures are two of the fastest models tested by \cite{tf_obj_det}.

\item[Faster-RCNN] As with SSD, We follow the implementation of \cite{tf_obj_det} (which closely follows \cite{faster_rcnn}), and adopt the ResNet 101 \cite{resnet101} architecture, (which \cite{tf_obj_det} noted as one of the ``sweet spot'' models in their comparison of speed/accuracy tradeoffs). 
We use the ``high-resolution" setting of $600 \times 600$ pixel image sizes, and use a batch size of 1 with an initial learning rate of  0.0001. 

\item[R-FCN] As with Faster-RCNN and SSD, we leverage the detailed optimization of \cite{tf_obj_det} for hyperparameter selection.  We utilize the ResNet 101 \cite{resnet101} architecture.  As with Faster-RCNN, we also explored the ResNet 50 architecture, but found no significant performance increase.  
We use the same parameters as Faster-RCNN, namely the ``high-resolution" setting of $600 \times 600$ pixel image sizes, and a batch size of 1. 


\end{description}

\section{Training Data}\label{sec_train}

\begin{figure}
\begin{center}
\includegraphics[height=6.5cm]{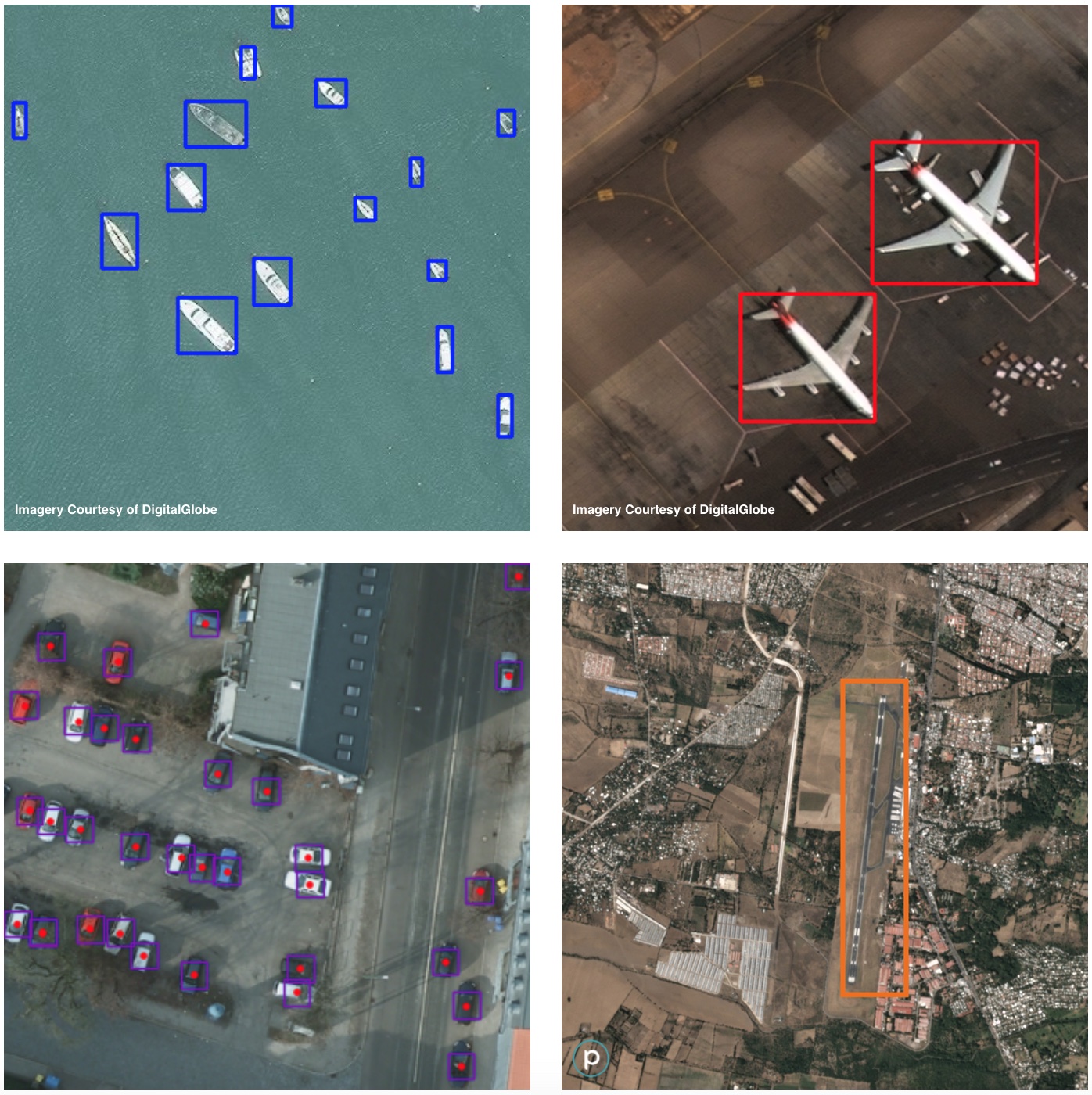}
\end{center}
\caption{SIMRDWN Training data examples. {\it Top left}: boats in DigitalGlobe imagery. {\it Top right}: airplanes in DigitalGlobe imagery.  {\it Bottom left}:  Cars from COWC \cite{cowc} aerial imagery; the red dow denotes the COWC label and the purple box is our inferred 3 meter bounding box. {\it Bottom right}: Airport labeled in Planet imagery.  }
\label{fig:SIMRDWN_training}
\end{figure}

Training data is collected from small chips of large images from three sources: DigitalGlobe satellites, Planet satellites, and aerial platforms. 
Labels are comprised of a bounding box and category identifier for each object (see Figure \ref{fig:SIMRDWN_training}).  
We initially focus on four categories: airplanes, boats, cars, and airports. 
For objects of very different scales (e.g.\ airplanes vs airports) we show in Section \ref{scale_conf_mit}  that using two different detectors at different scales is very effective.

\begin{description}
\item[Cars] The Cars Overhead with Context (COWC) \cite{cowc} dataset is a large, high quality set of annotated cars from overhead imagery collected over multiple locales.  Data is collected via aerial platforms, but at a nadir view angle such that it resembles satellite imagery.
The  imagery has a resolution of 15 cm GSD that is approximately double the current best resolution of commercial satellite imagery (30 cm GSD for DigitalGlobe).  Accordingly, we convolve the raw imagery with a Gaussian kernel and reduce the image dimensions by half to create the equivalent of 30 cm GSD images.  Labels consist of simply a dot at the centroid of each car, and we draw a 3 meter bounding box around each car for training purposes.  We reserve the largest geographic region (Utah) for testing, leaving 13,303 labeled training cars.  Training images are cut into 416 pixel chips, corresponding to 125 meter window sizes. 


\item[Airplanes] We labeled eight DigitalGlobe images over airports for a total of 230 objects in the training set.    Training images are cut into chips of 125-200 meters depending on resolution.

\item[Boats] We labeled three DigitalGlobe images taken over coastal regions for a total of 556 boats.  Training images are cut into chips of 125-200 meters depending on resolution.

\item[Airports] We labeled airports in 37 Planet images for training purposes, each with a single airport per 5000m chip. Obviously, the lower resolution of Planet imagery of 3-4 meter GSD limits the utility of this imagery for vehicle detection.  

\end{description}

To address unusual aspect ratios and configurations we augment this training data by rotating training images about the unit circle to ensure that the classifier is agnostic to object heading.  We also randomly scale the images in HSV (hue-saturation-value) to increase the robustness of the classifier to varying sensors, atmospheric conditions, and lighting conditions.  Even with augmentation, the raw training datasets for airplanes, airports, and watercraft are quite small by computer vision standards, and a larger dataset may improve the inference performance detailed in Section~\ref{sec_results}. 

A potential additional data source is provided by the SpaceNet satellite imagery dataset\footnote{https://spacenetchallenge.github.io/}, which contains a large corpus of labeled building footprints in polygon (not bounding box) format.  While bounding boxes are not ideal for precise building footprint estimation, this dataset nevertheless merits future investigation.  The impending release of the X-View satellite imagery dataset \cite{xview} with 60 object classes and approximately one million labeled object instances will also be of great use for training purposes once available.



\section{Test Images}\label{sec_test}

To ensure test robustness and to penalize overtraining on background features, all test images are taken from different geographic regions than training examples.  Our dataset for airports is the smallest, with only ten Planet images available for testing.  See Table~\ref{tab_train_test_split} for the train/test split for each category. For airplane testing we label four DigitalGlobe images for a total of 74 airplanes.  Our airplane training dataset contains only airliners, though some of the test object are small personal aircraft -not all classifiers perform well on these objects.  Two DigitalGlobe and two Planet coastal images are labeled, yielding 771 test boats.  Since we extract test objects from different images than our training set, the sea state is also different in our test images compared to the training images.  In addition, two of the four coastal test images are from 3 meter resolution Planet imagery, which further tests the robustness of our models since all training objects are taken from high resolution 0.30 or 0.50 meter DigitalGlobe imagery.  
The externally labeled cars test dataset is by far the largest; we reserve the largest geographic region of the COWC dataset (Utah) for testing, yielding 19,807 test cars. 

\begin{table}
\begin{center}
\begin{threeparttable}
\caption{Train/Test Split} 
\label{tab_train_test_split}
\begin{tabular}{l c c}
Object Class\;	& Training Examples\; & 	Test Examples  \\
\hline
Airport\tnote{$\ast$}		& 37 &		10 \\
Airplane\tnote{$\ast$}	& 230 &	 	74\\
Boat\tnote{$\ast$}		& 556 &	  	100 \\
\hline
Car\tnote{$\dagger$}		& 13,303 &	19,807 \\
\end{tabular}
\begin{tablenotes}
\item[$\ast$] Internally labeled
\item[$\dagger$] External Dataset
\end{tablenotes}
\end{threeparttable}
\end{center}
\end{table}

\section{Experiment Procedure}\label{sec_test_proc}

\subsection{Training}
Each of the five architectures discussed in Section \ref{sec_arch} (Faster RCNN Resnet 101, R-FCN Resnet 101, SSD Inception v2, SSD MobileNet, YOLT) are trained on the same data.  We create a list of training images and labels for YOLT training, and transform that list into a tfrecord for training the tensorflow models.  Models are trained for approximately the same amount of time (as detailed in Section \ref{sec_arch}) of 24-48 hours. We train two separate models for each architecture, one designed for vehicles, and the other for airports (for the rationale behind this approach, see Section \ref{sec_res_prelim}).

\subsection{Test Evaluation}\label{sec_eval}

For each test image we execute Sections \ref{sec_infer} and \ref{sec_post_proc} to yield bounding box predictions.
For comparison of predictions to ground truth we define a true positive as having an intersection over union (IOU) of greater than a given threshold. An IoU of 0.5 is often used as the threshold for a correct detection.  We adopt an IoU of 0.5 to indicate a true positive, though we adopt a lower threshold for cars (which typically only 10 pixels in extent) of 0.25.  This mimics Equation 5 of ImageNet \cite{imagenet}, which sets an IoU threshold of 0.25 for objects 10 pixels in extent.  



Precision-recall curves are computed by evaluating test images over a range of probability thresholds.  At each of 30 evenly spaced thresholds between 0.05 and 0.95, we discard all detections below the given threshold.  Non-max suppression for each object class is subsequently applied to the remaining bounding boxes; the precision and recall at that threshold is tabulated from the summed true positive, false positive, and false negatives of all test images.  Finally, we compute the average precision (AP) for each object class and each model, along with the mean average precision (mAP) for each model.

\section{Object Detection Results}\label{sec_results}

\subsection{Preliminary Object Detection Results}\label{sec_res_prelim}

\begin{figure}
\begin{center}
\includegraphics[height=5cm]{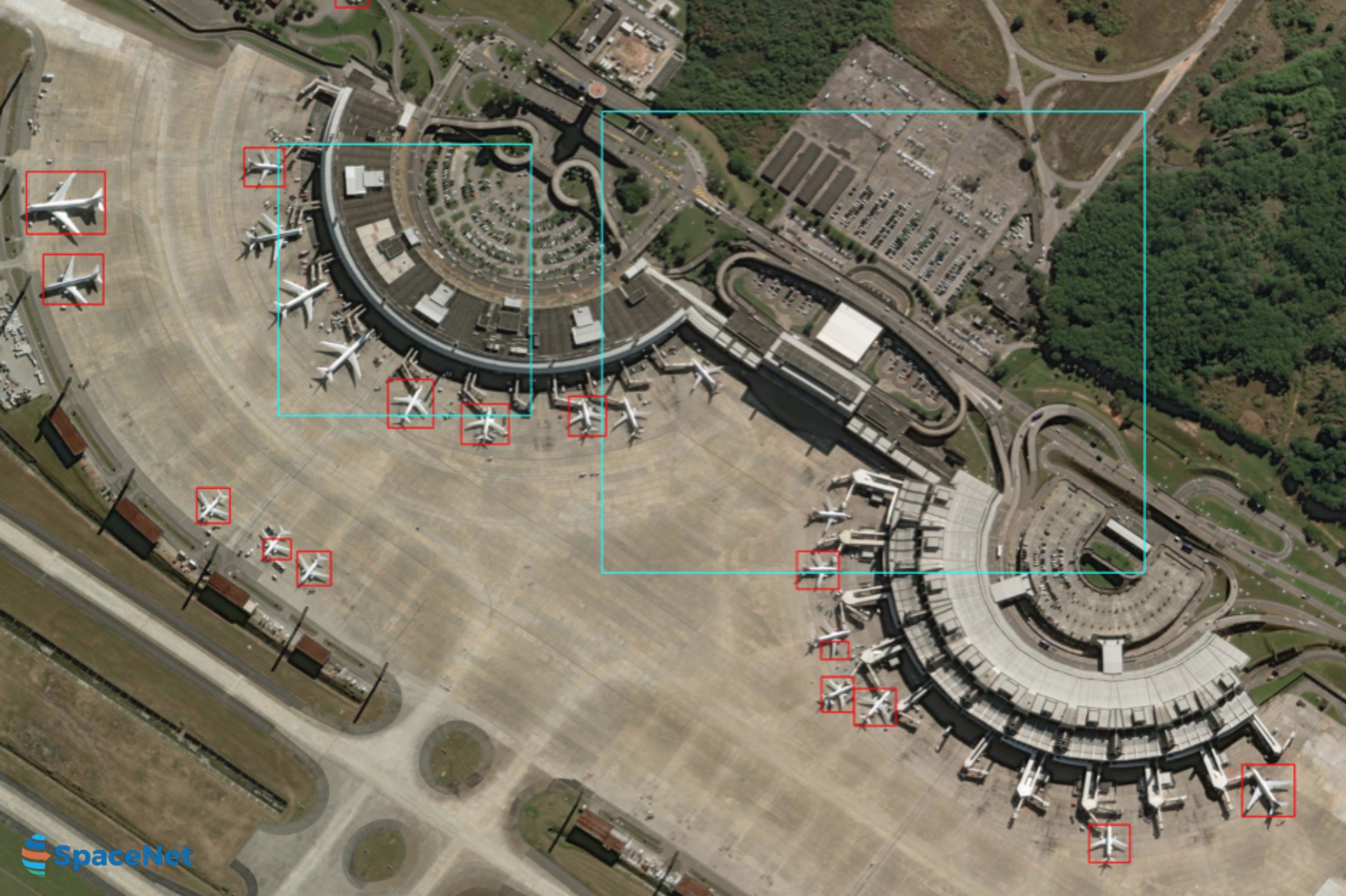}
\end{center}
\caption{Poor results of the universal YOLT model applied to DigitalGlobe imagery on two different scales (200m, 1500m). Airplanes are in red. The cyan boxes mark spurious detections of runways, caused in part by confusion from small scale linear structures such as highways.}
\label{fig:airport_spurious}
\end{figure}

Initially, we attempt to train a single classifier to recognize all four categories listed above, both vehicles and airports.  We note a number of spurious airport detections in this example (see Figure \ref{fig:airport_spurious}), as downsampled runways look similar to highways at the wrong scale.  

\subsection{Scale Confusion Mitigation} \label{scale_conf_mit}

There are multiple ways one could address the false positive issues noted in Figure \ref{fig:airport_spurious}.  Recall from Section  \ref{sec_train} that for this exploratory work our training set consists of only a few dozen airports, far smaller than usual for deep learning models.  Increasing this training set size could greatly improve our model, particularly if the background is highly varied.  Another option would involve a post-processing step to remove any detections at the incorrect scale (e.g.\ an airport with a size of $\sim50$ meters).  Another option is to simply build dual classifiers, one for each relevant scale. 

We opt to utilize the scale information present in satellite imagery and run two different classifiers: one trained for vehicles/buildings, and the other trained only to look for airports in downsampled Planet images a few kilometers in extent.  Running a second classifier at a larger scale has a negligible impact on runtime performance, since in a given image there are $\approx1\%$ as many 2000 meter chips as 200 meter chips. 

\subsection{Results} \label{sec_vehicle}

\begin{figure}
\begin{center}
\includegraphics[height=6.5cm]{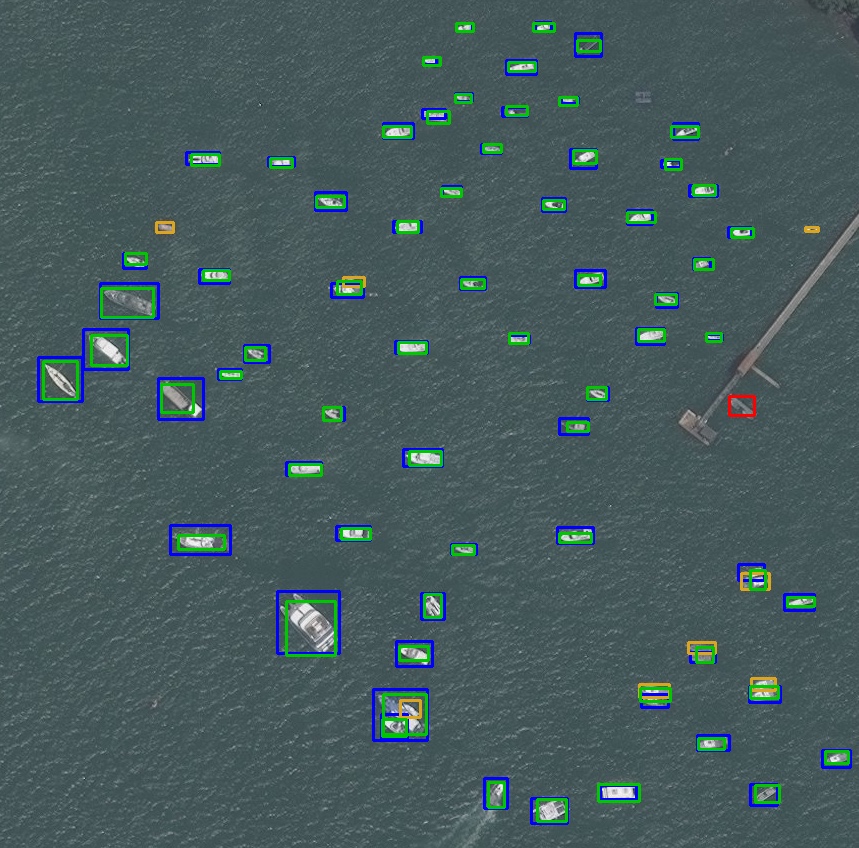} 
\end{center}
\caption{Portion of evaluation image with the YOLT model showing labeled boats.  False positives are shown in red, false negatives are yellow, true positives are green, and blue rectangles denote ground truth for all true positive detections.  
}
\label{fig:fig_yolt_panama_sample}
\end{figure}

\begin{figure}
\begin{center}
\includegraphics[width=0.95\linewidth]{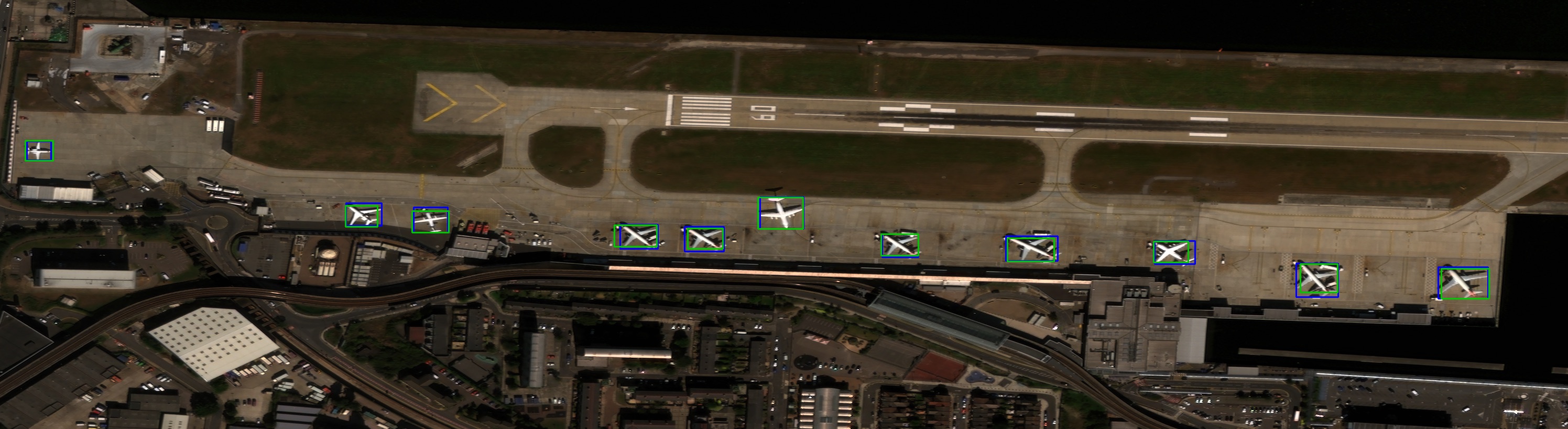} 
\end{center}
\caption{Portion of evaluation image with the YOLT model showing labeled aircraft.  False positives are shown in red, false negatives are yellow, true positives are green, and blue rectangles denote ground truth for all true positive detections.  Performance is good despite the atypical look angle and lighting conditions.
}
\label{fig:fig_yolt_rio_planes}
\end{figure}

\begin{figure*}
\begin{center}
\begin{tabular}{lcr}
  \includegraphics[width=55mm]{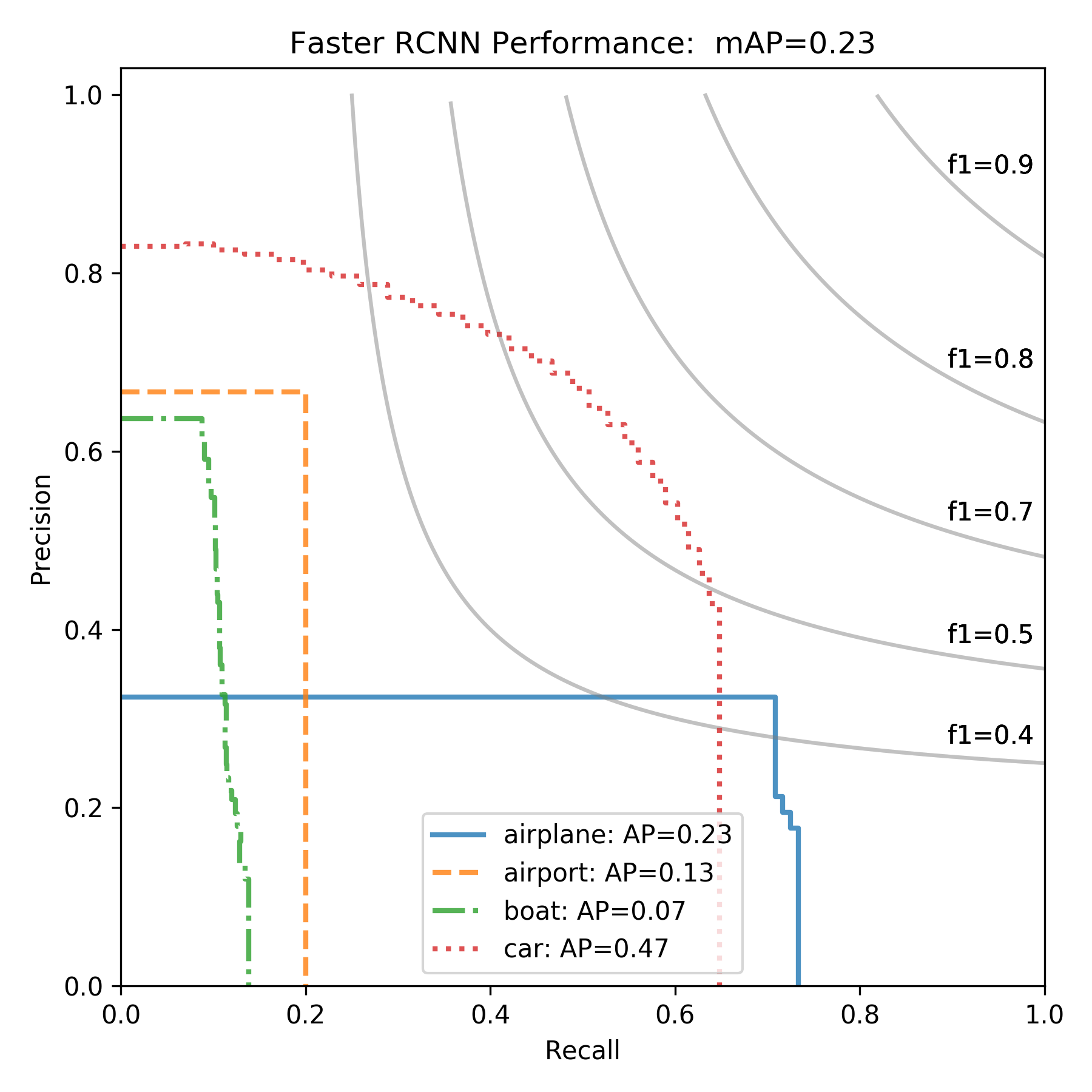} &   \includegraphics[width=55mm]{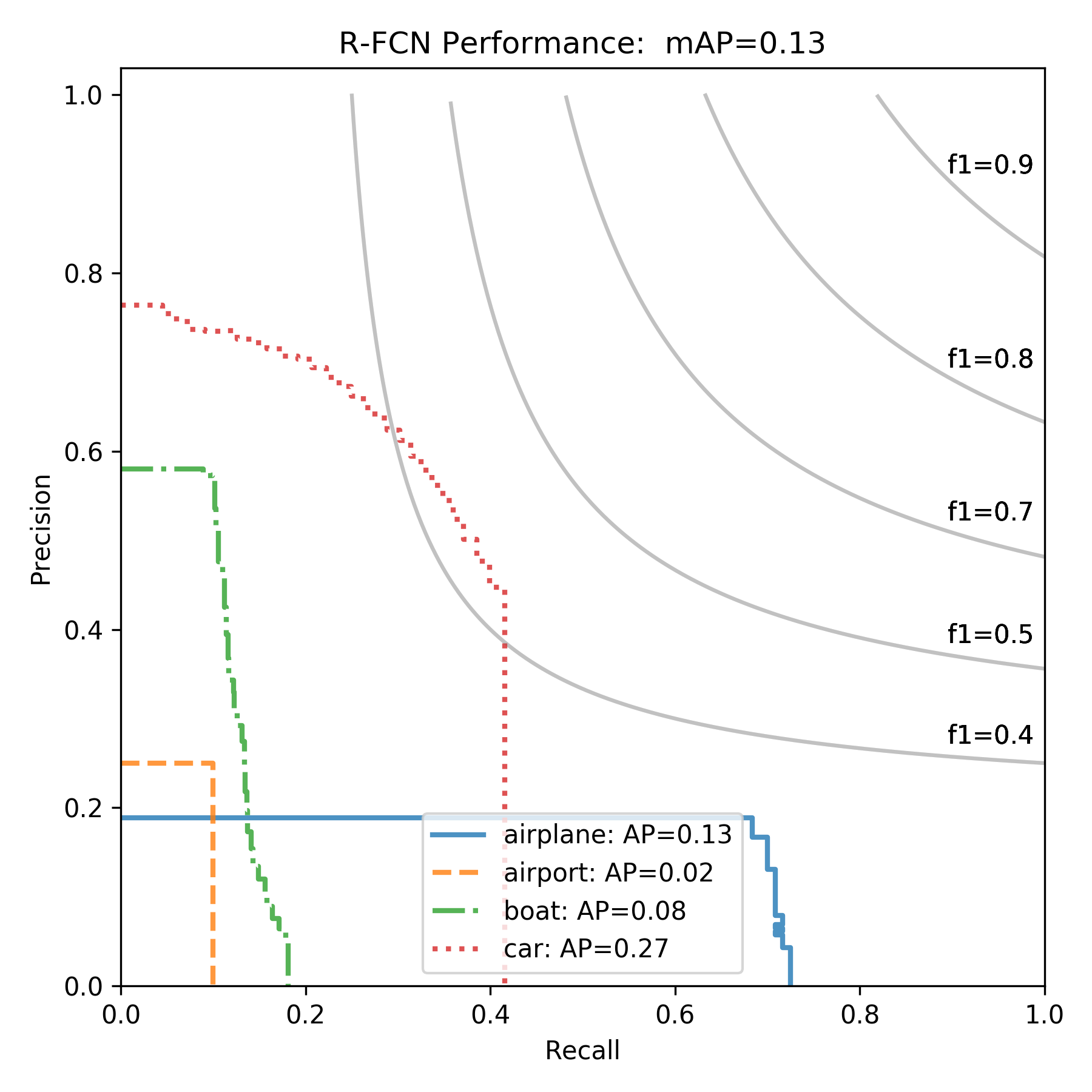} &  \includegraphics[width=55mm]{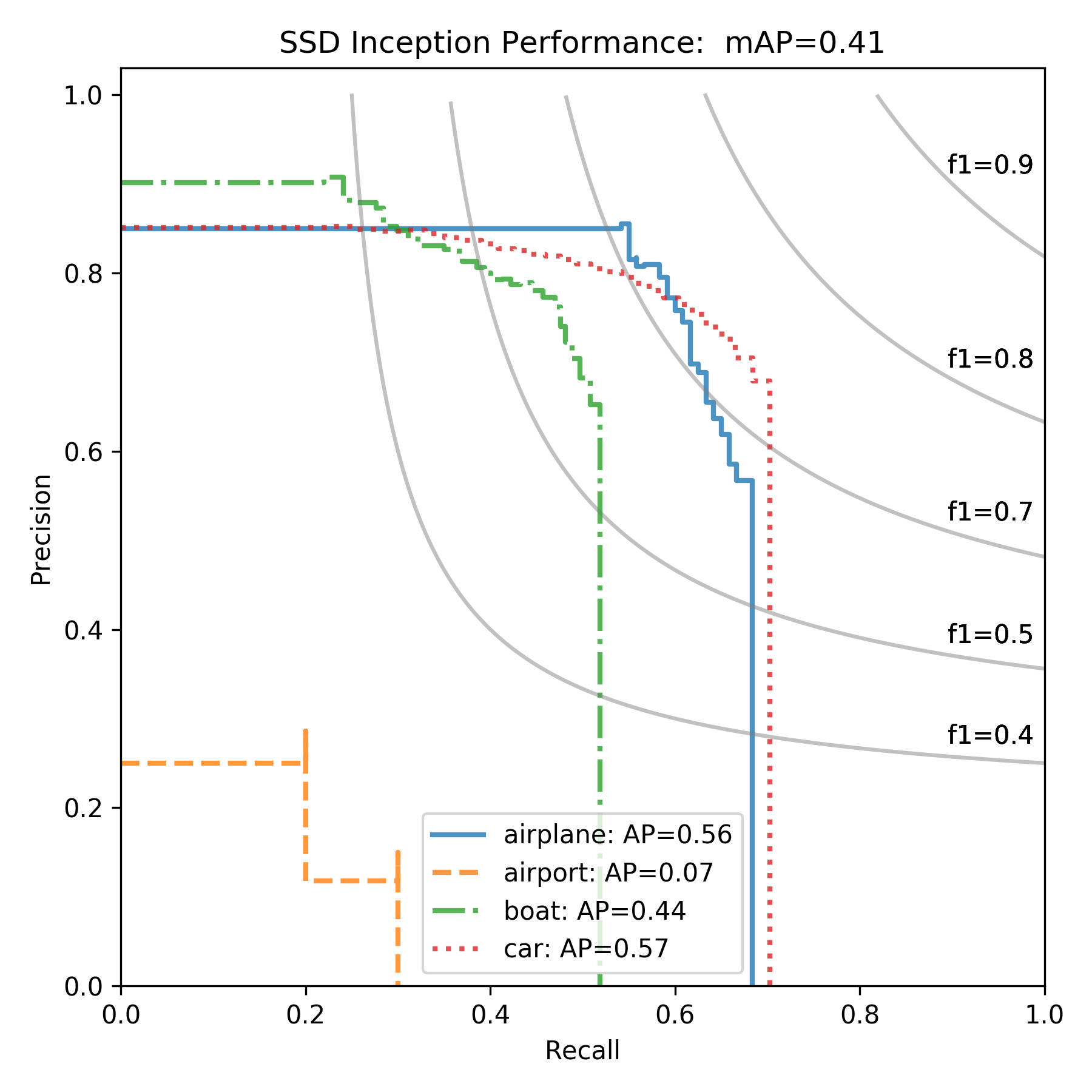} \\
 \includegraphics[width=55mm]{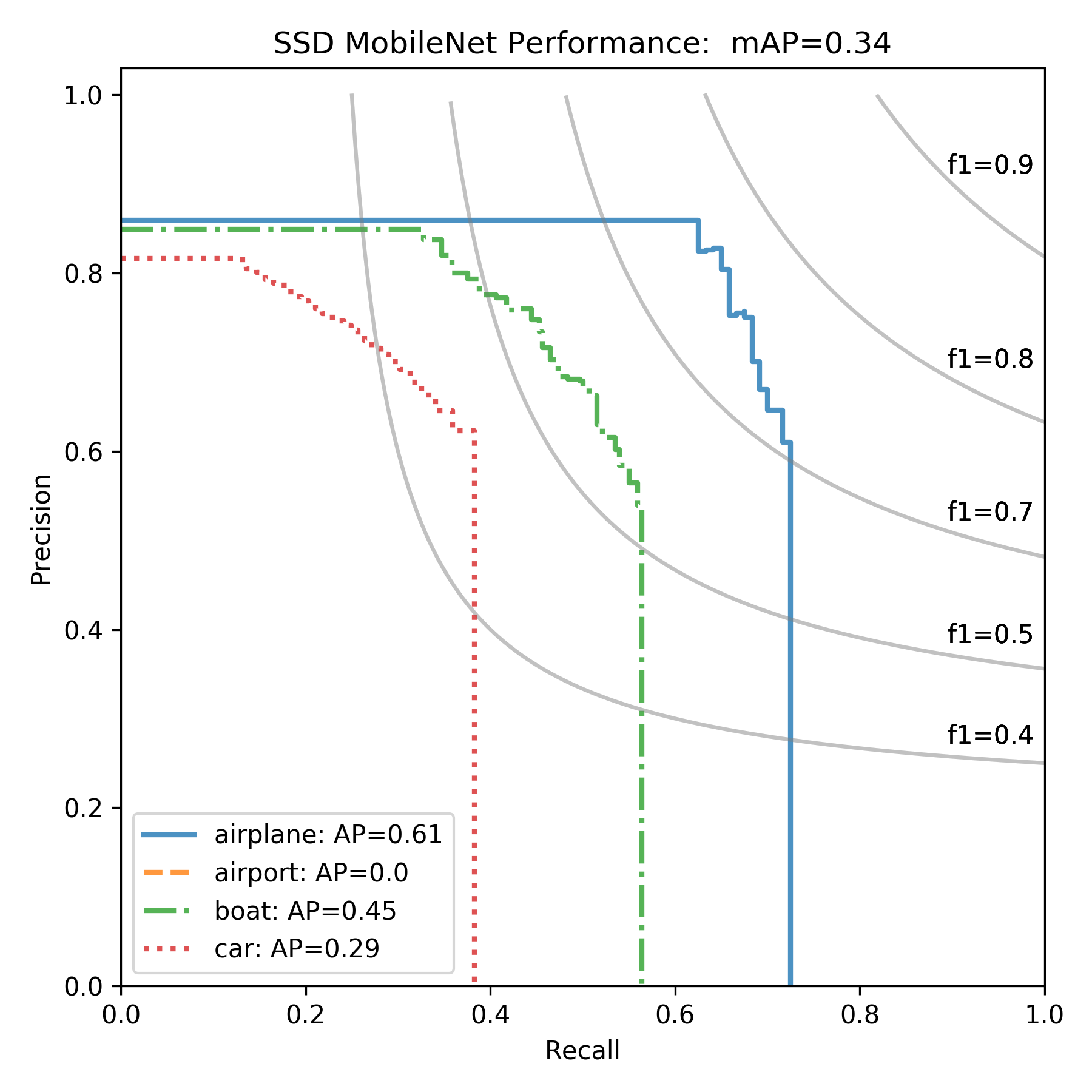} & \includegraphics[width=55mm]{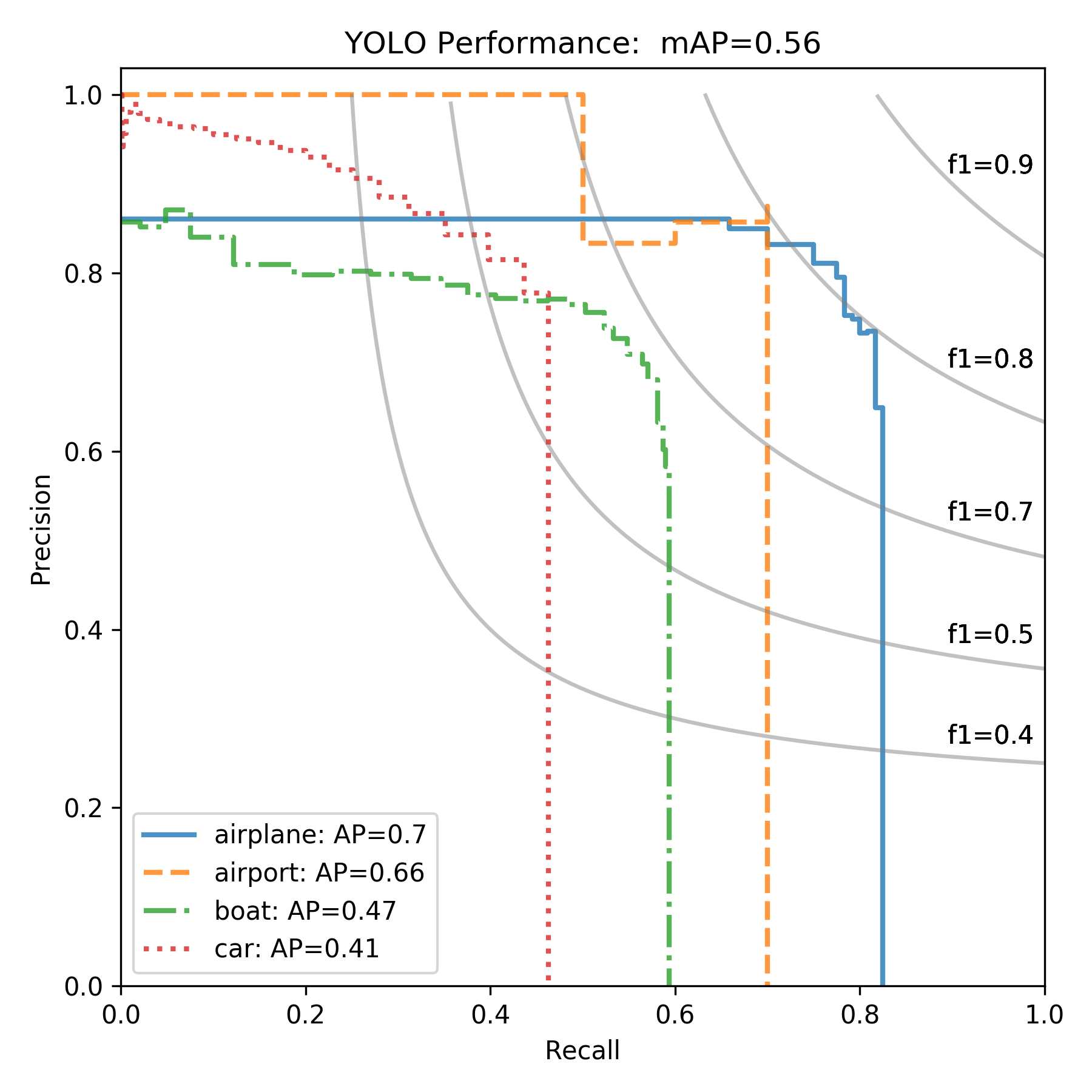} &   \includegraphics[width=55mm]{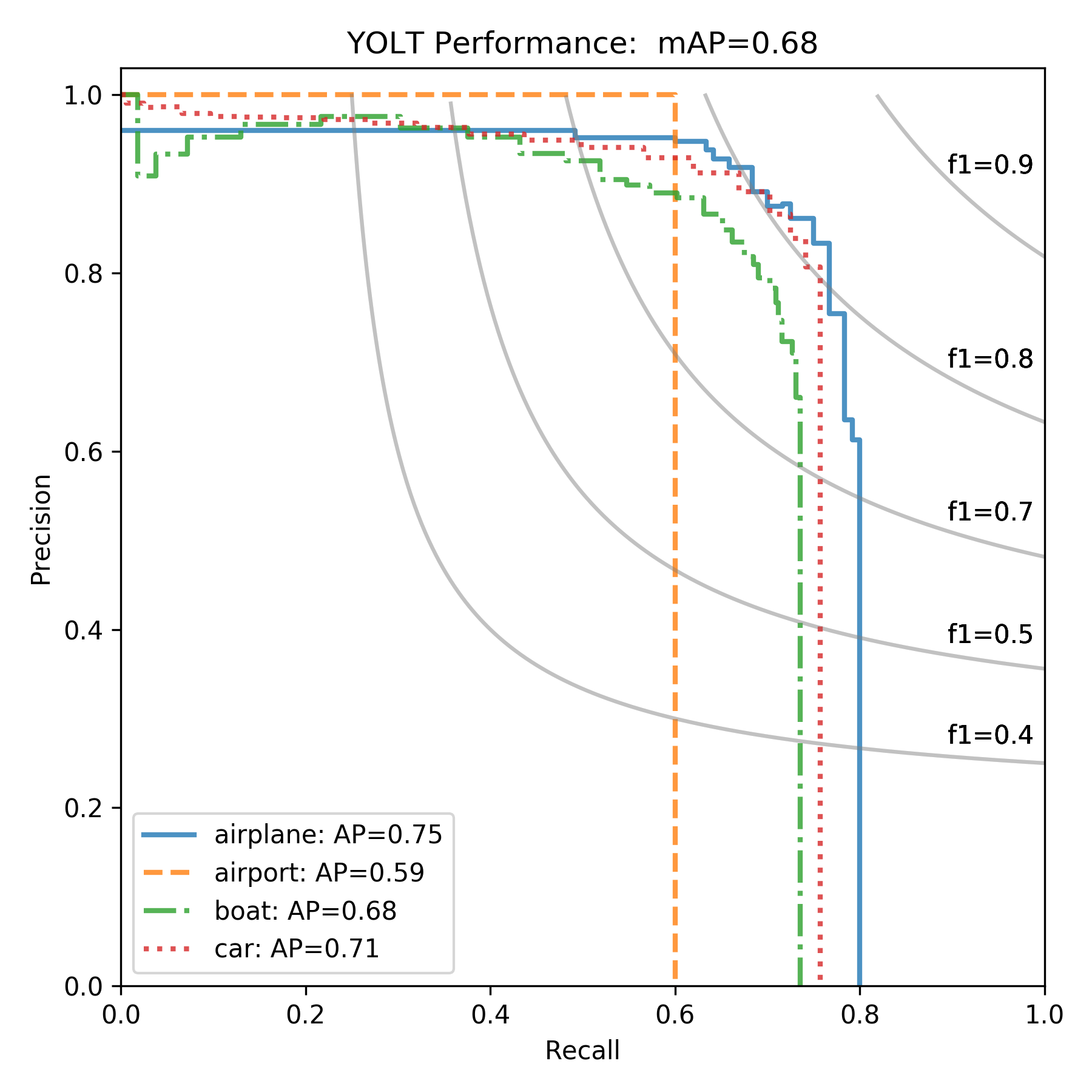} \\
\end{tabular}
\caption{Precision-recall curves for each model}
\label{fig:fig_perf}
\end{center}
\end{figure*}

For large validation images, we run the classifier at two different scales: 200m, and 5000m. The first scale is designed for vehicles (see Figures \ref{fig:fig_yolt_panama_sample}, \ref{fig:fig_yolt_rio_planes}), and the larger scale is optimized for large infrastructure such as airports (see Supplemental Material). We break the validation image into appropriately sized bins and run each image chip on the appropriate classifier. The myriad results from the many image chips and multiple classifiers are combined into one final image, and overlapping detections are merged via non-maximal suppression. Model performance is shown in Figure \ref{fig:fig_perf}.

%
%

Results for R-FCN and Faster-RCNN do not track with the conclusions of \cite{tf_obj_det} that these models occupy a ``sweet spot'' in terms of speed and accuracy.  Inspection of results indicate that both these models struggle with different object sizes (e.g.\ boats larger than the typical training example), and are very sensitive to background conditions.  In an effort to improve results, we experiment with model hyperparameters, and for each model we we explore the following: training runs of [30,000, 100,000, 300,000] iterations, input image size of [416, 600], first stage stride of [8, 16], batch size of [1, 4, 8].  These experiments yield no improvement over the default hyperparameters listed in Section \ref{sec_arch}, so it appears that at least for our dataset Faster RCNN and R-FCN struggle to localize objects of interest.  

Airport detection is poor for all models, likely the result of the small training set size, since airports are a large and distinctive feature that do not suffer from many of the complications listed in Section \ref{sec_intro}.  It does appear that the YOLO/YOLT models perform significantly better with this  training set, though further research is required to determine if these models are truly more robust or if another mechanism explains the superior performance of YOLO/YOLT to other models.  We also note a significant increase in mAP from YOLO to YOLT,  which stems from improved localization of cars and boats (which are often tightly packed) where the denser network of YOLT pays dividends.

Table \ref{tab_f1_speed} displays object detection performance and speed for each model architecture.  
We report inference speed in terms of GPU time to run the inference step.  
Currently, pre-processing (i.e.\ splitting test images into smaller cutouts) and post-processing (i.e.\ stitching results back into one global image) is not fully optimized and is performed on the CPU, which increases run time by a factor of $1.5 - 1.75$.  Inference rates for airports are $\sim600\times$ faster than the inference rate for vehicles reported in Table \ref{tab_f1_speed}, ranging from 60 $\rm{km^2/s}$ (Faster RCNN) to 270 $\rm{km^2/s}$ (YOLT).

\setlength{\tabcolsep}{4pt}
\begin{table}
\begin{center}
\begin{threeparttable}
\caption{Performance vs Speed} 
\label{tab_f1_speed}
\begin{tabular}{l c c}
Architecture	& mAP & 	Inference Rate \\
& &  ($\rm{km^2/s}$) \\
\hline

Faster RCNN ResNet101		& $0.23$ & 0.09 \\
RFCN ResNet101			& $0.13$ & 0.17 \\
SSD Inception				& $0.41$ & 0.22 \\
SSD MobileNet				& $0.34$ & 0.32\\
YOLO					& $0.56$ & 0.42\\
YOLT					& $\textbf{0.68}$ & $\textbf{0.44}$\\


\end{tabular}
\end{threeparttable}
\end{center}
\end{table}

\subsection{Resolution Performance}

\begin{figure}
\begin{center}
\includegraphics[height=55mm]{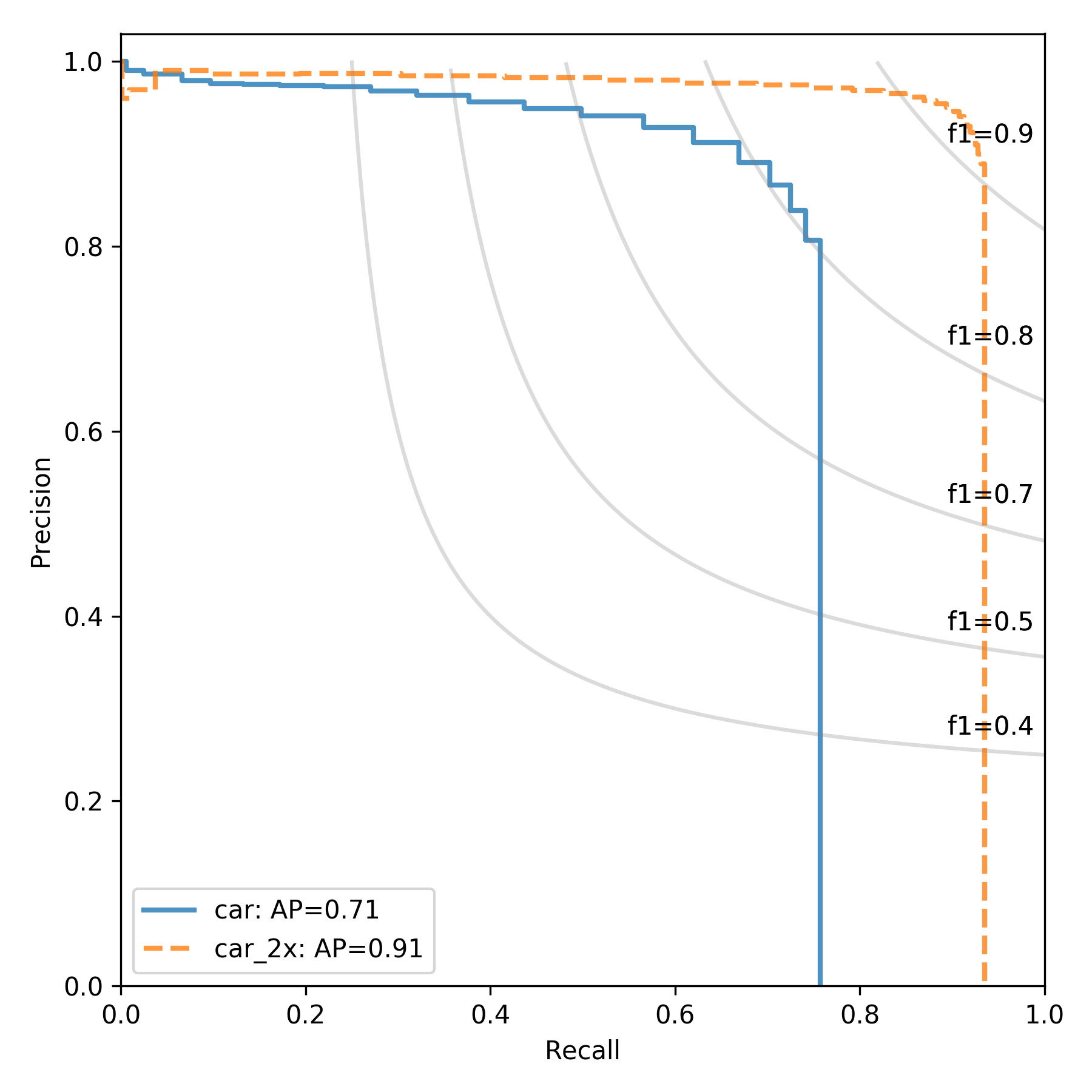} 
\end{center}
\caption{Performance of the YOLT model trained and tested at native resolution (solid), and a model trained and tested at simulated double resolution (dashed).}
\label{fig:fig_2x}
\end{figure}

We explore the effect of window size (closely related to image resolution) on object detection performance.  The YOLT model returns the best AP for cars, though dense regions still pose a challenge for the detector.  The YOLT model is trained on native resolution imagery of 416 pixels in extent. In an attempt to improve performance, we train on image cutouts of only 208 pixels, these cutouts are subsequently upsampled to size 416 pixels when ingested by the network.  This simulates higher resolution imagery, though no extra information is provided.  This smaller window size decreases inference speed by a factor of four, but markedly improves performance, see Figure \ref{fig:fig_2x}.
 
\section{Conclusions}
Object detection algorithms have made great progress as of late in localizing objects in 
ImageNet style datasets.  Such algorithms are rarely well suited to the object sizes or orientations present in satellite imagery, however, nor are they designed to handle images with hundreds of megapixels.  

To address these limitations we implemented a fully convolutional neural network pipeline (SIMRDWN) to rapidly localize vehicles and airports in satellite imagery.  This pipeline unifies leading object detection algorithms such as SSD, Faster RCNN, R-FCN, and YOLT into a single framework that rapidly analyzes test images of arbitrary size.  We noted poor results from a combined classifier due to confusion between small and large features, such as highways and runways.  Training dual classifiers at different scales (one for vehicles, and one for airports), yielded far better results.

Our training dataset is quite small by computer vision standards, and mAP scores range from 0.13 (R-FCN) to 0.68 (YOLT) for our test set.   While the mAP scores may not be at the level many readers are accustomed to from ImageNet competitions, object detection in satellite imagery is still a relatively nascent field and has unique challenges. In addition, our training dataset for most categories is relatively small for supervised learning methods.

 Our test set is derived from different geographic regions than the training set, and the low mAP scores are unsurprising given the small training set size provides relatively little background variation.  Nevertheless, the YOLT architecture did perform significantly better than the other rapid object detection frameworks, indicating that it appears better able to disentangle objects from background with small training sets.  

Inference speeds for vehicles are high, at 0.09 $\rm{km^2/s}$ (Faster RCNN) to 0.44 $\rm{km^2/s}$ (YOLT). 
We also demonstrated the ability to train on one sensor (e.g.\ DigitalGlobe), and apply our model to a different sensor (e.g.\ Planet).  
The highest inference speed translates to a runtime of $<6$ minutes to localize all vehicles in an area of the size of Washington DC, and $<2$ seconds to localize airports over this area.  DigitalGlobe's WorldView3 
satellite\footnote{http://worldview3.digitalglobe.com} 
covers a maximum of 680,000 km$^2$ per day, so at SIMRDWN inference speed a 16 GPU cluster would provide real-time inference on satellite imagery.
Results so far are intriguing, and it will be interesting to explore in future works how well the SIMRDWN pipeline performs as further datasets become available and the number of object categories increases.

{\small
\bibliographystyle{ieee}
\bibliography{bib}

\begin{thebibliography}{10}\itemsep=-1pt

\bibitem{people-art_dataset}
H.~Cai, Q.~Wu, T.~Corradi, and P.~Hall.
\newblock The cross-depiction problem: Computer vision algorithms for
  recognising objects in artwork and in photographs.
\newblock {\em CoRR}, abs/1505.00110, 2015.

\bibitem{rs_obj_det_survey}
G.~Cheng and J.~Han.
\newblock A survey on object detection in optical remote sensing images.
\newblock {\em CoRR}, abs/1603.06201, 2016.

\bibitem{rfcn}
J.~Dai, Y.~Li, K.~He, and J.~Sun.
\newblock {R-FCN:} object detection via region-based fully convolutional
  networks.
\newblock {\em CoRR}, abs/1605.06409, 2016.

\bibitem{pascal_voc}
M.~Everingham, L.~Van~Gool, C.~K.~I. Williams, J.~Winn, and A.~Zisserman.
\newblock The pascal visual object classes (voc) challenge.
\newblock {\em International Journal of Computer Vision}, 88(2):303--338, June
  2010.

\bibitem{picasso_dataset}
S.~Ginosar, D.~Haas, T.~Brown, and J.~Malik.
\newblock Detecting people in cubist art.
\newblock {\em CoRR}, abs/1409.6235, 2014.

\bibitem{resnet}
K.~He, X.~Zhang, S.~Ren, and J.~Sun.
\newblock Deep residual learning for image recognition.
\newblock {\em CoRR}, abs/1512.03385, 2015.

\bibitem{resnet101}
K.~He, X.~Zhang, S.~Ren, and J.~Sun.
\newblock Deep residual learning for image recognition.
\newblock {\em CoRR}, abs/1512.03385, 2015.

\bibitem{mobilenet}
A.~G. Howard, M.~Zhu, B.~Chen, D.~Kalenichenko, W.~Wang, T.~Weyand,
  M.~Andreetto, and H.~Adam.
\newblock Mobilenets: Efficient convolutional neural networks for mobile vision
  applications.
\newblock {\em CoRR}, abs/1704.04861, 2017.

\bibitem{tf_obj_det}
J.~Huang, V.~Rathod, C.~Sun, M.~Zhu, A.~Korattikara, A.~Fathi, I.~Fischer,
  Z.~Wojna, Y.~Song, S.~Guadarrama, and K.~Murphy.
\newblock Speed/accuracy trade-offs for modern convolutional object detectors.
\newblock {\em CoRR}, abs/1611.10012, 2016.

\bibitem{inceptionv2}
S.~Ioffe and C.~Szegedy.
\newblock Batch normalization: Accelerating deep network training by reducing
  internal covariate shift.
\newblock {\em CoRR}, abs/1502.03167, 2015.

\bibitem{alexnet}
A.~Krizhevsky, I.~Sutskever, and G.~E. Hinton.
\newblock Imagenet classification with deep convolutional neural networks.
\newblock In F.~Pereira, C.~J.~C. Burges, L.~Bottou, and K.~Q. Weinberger,
  editors, {\em Advances in Neural Information Processing Systems 25}, pages
  1097--1105. Curran Associates, Inc., 2012.

\bibitem{xview}
D.~Lam, R.~Kuzma, K.~McGee, S.~Dooley, M.~Laielli, M.~Klaric, Y.~Bulatov, and
  B.~McCord.
\newblock xview: Objects in context in overhead imagery.
\newblock {\em CoRR}, abs/1802.07856, 2018.

\bibitem{ms_coco}
T.~Lin, M.~Maire, S.~J. Belongie, L.~D. Bourdev, R.~B. Girshick, J.~Hays,
  P.~Perona, D.~Ramanan, P.~Doll{\'{a}}r, and C.~L. Zitnick.
\newblock Microsoft {COCO:} common objects in context.
\newblock {\em CoRR}, abs/1405.0312, 2014.

\bibitem{ssd}
W.~Liu, D.~Anguelov, D.~Erhan, C.~Szegedy, S.~E. Reed, C.~Fu, and A.~C. Berg.
\newblock {SSD:} single shot multibox detector.
\newblock {\em CoRR}, abs/1512.02325, 2015.

\bibitem{rs_obj_det_slow}
Y.~Long, Y.~Gong, Z.~Xiao, and Q.~Liu.
\newblock Accurate object localization in remote sensing images based on
  convolutional neural networks.
\newblock {\em IEEE Transactions on Geoscience and Remote Sensing},
  55(5):2486--2498, May 2017.

\bibitem{samfinder}
R.~A. {Marcum}, C.~H. {Davis}, G.~J. {Scott}, and T.~W. {Nivin}.
\newblock {Rapid broad area search and detection of Chinese surface-to-air
  missile sites using deep convolutional neural networks}.
\newblock {\em Journal of Applied Remote Sensing}, 11(4):042614, Oct. 2017.

\bibitem{mnihroads}
V.~Mnih and G.~E. Hinton.
\newblock Learning to detect roads in high-resolution aerial images.
\newblock In K.~Daniilidis, P.~Maragos, and N.~Paragios, editors, {\em Computer
  Vision -- ECCV 2010}, pages 210--223, Berlin, Heidelberg, 2010. Springer
  Berlin Heidelberg.

\bibitem{cowc}
T.~N. Mundhenk, G.~Konjevod, W.~A. Sakla, and K.~Boakye.
\newblock A large contextual dataset for classification, detection and counting
  of cars with deep learning.
\newblock {\em CoRR}, abs/1609.04453, 2016.

\bibitem{darknet}
J.~Redmon.
\newblock Darknet: Open source neural networks in c.
\newblock {\em http://pjreddie.com/darknet/}, 2013-2017.

\bibitem{yolov1}
J.~Redmon, S.~K. Divvala, R.~B. Girshick, and A.~Farhadi.
\newblock You only look once: Unified, real-time object detection.
\newblock {\em CoRR}, abs/1506.02640, 2015.

\bibitem{yolo9000}
J.~Redmon and A.~Farhadi.
\newblock {YOLO9000:} better, faster, stronger.
\newblock {\em CoRR}, abs/1612.08242, 2016.

\bibitem{faster_rcnn}
S.~Ren, K.~He, R.~B. Girshick, and J.~Sun.
\newblock Faster {R-CNN:} towards real-time object detection with region
  proposal networks.
\newblock {\em CoRR}, abs/1506.01497, 2015.

\bibitem{imagenet}
O.~Russakovsky, J.~Deng, H.~Su, J.~Krause, S.~Satheesh, S.~Ma, Z.~Huang,
  A.~Karpathy, A.~Khosla, M.~Bernstein, A.~C. Berg, and L.~Fei-Fei.
\newblock {ImageNet Large Scale Visual Recognition Challenge}.
\newblock {\em International Journal of Computer Vision (IJCV)},
  115(3):211--252, 2015.

\bibitem{vakabuildings}
M.~Vakalopoulou, K.~Karantzalos, N.~Komodakis, and N.~Paragios.
\newblock Building detection in very high resolution multispectral data with
  deep learning features.
\newblock In {\em 2015 IEEE International Geoscience and Remote Sensing
  Symposium (IGARSS)}, pages 1873--1876, July 2015.

\bibitem{yolt}
A.~{Van Etten}.
\newblock {You Only Look Twice: Rapid Multi-Scale Object Detection In Satellite
  Imagery}.
\newblock {\em ArXiv e-prints}, May 2018.

\bibitem{roadunet}
Z.~Zhang, Q.~Liu, and Y.~Wang.
\newblock Road extraction by deep residual u-net.
\newblock {\em CoRR}, abs/1711.10684, 2017.

\end{thebibliography}
}

\clearpage
\newpage


\appendix
\section{Image Appendix} \label{appendix}


\begin{figure}[h]
\begin{center}
\includegraphics[width=0.95\linewidth]{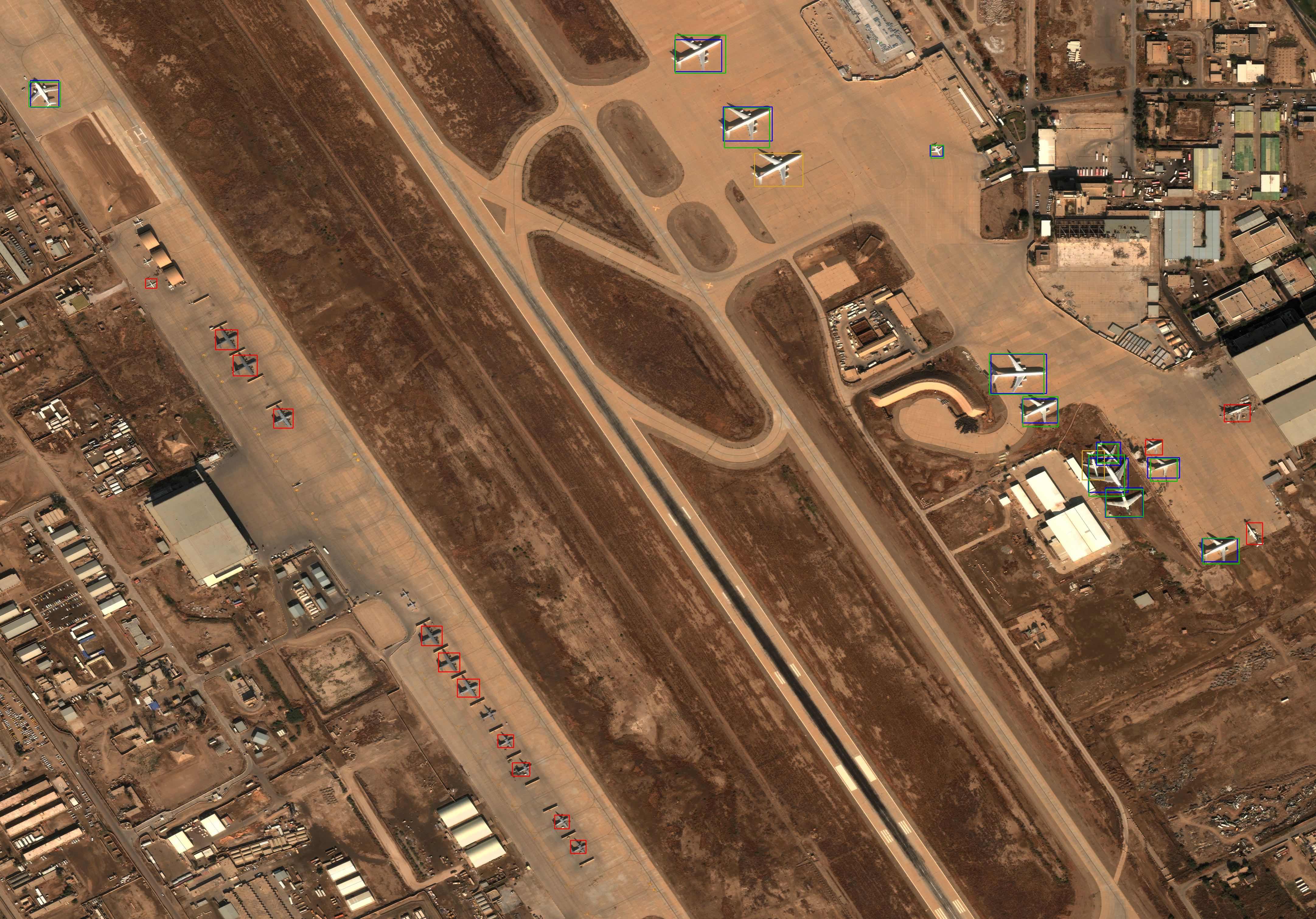} 
\end{center}
\caption{Portion of evaluation image with the YOLT model.  False positives are shown in red, false negatives are yellow, true positives are green, and blue rectangles denote ground truth for all true positive detections.  This image demonstrates some of the challenges of our test set and the robustness of the model. Our airplane training set only contains airliners, so we only label commercial aircraft in test images, yet the many false negatives in this image are caused by detections of military aircraft.
}
\label{fig:fig_cowc}
\end{figure}

\begin{figure}[h]
\begin{center}
\includegraphics[width=0.95\linewidth]{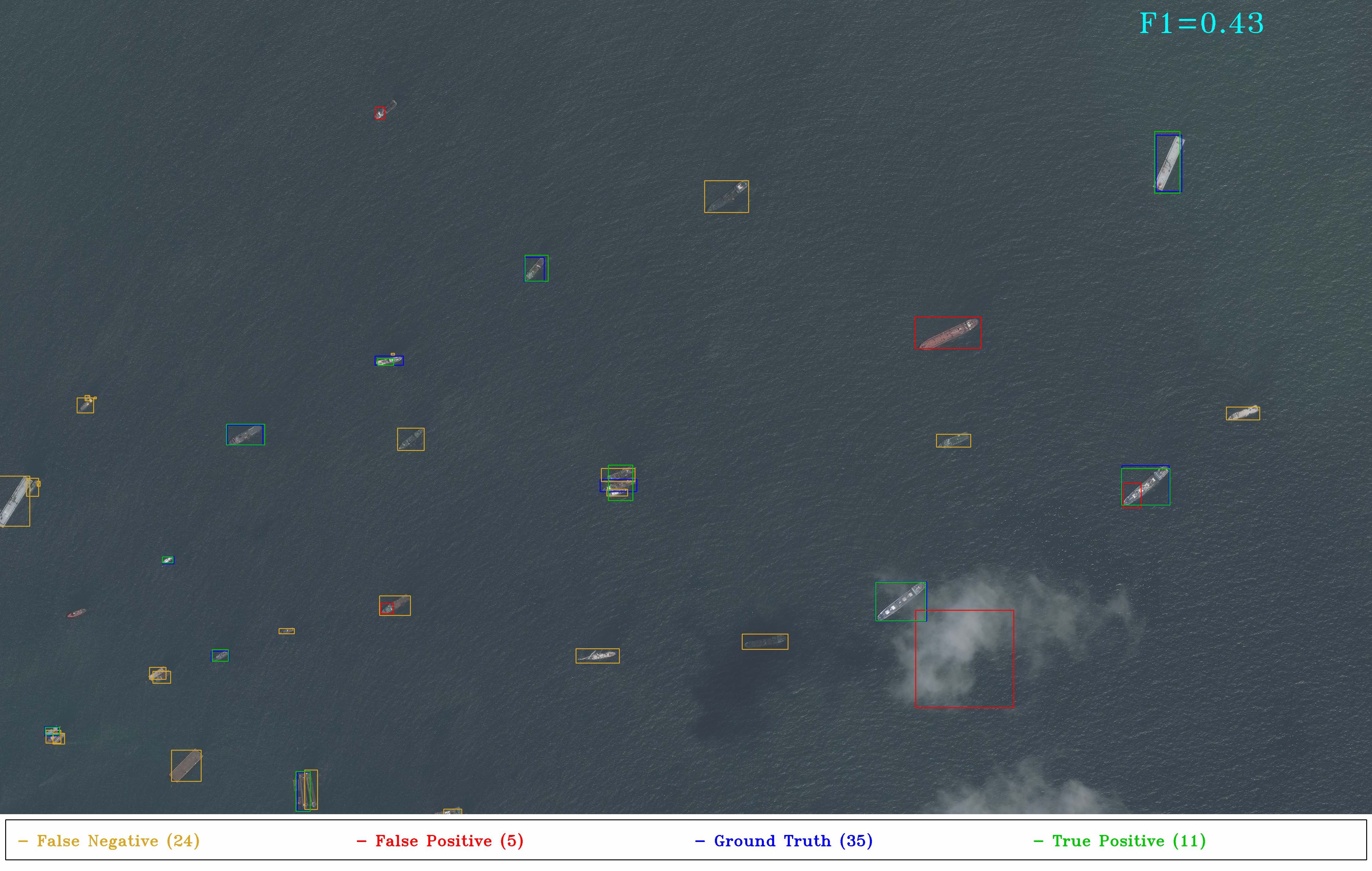} 
\end{center}
\caption{Evaluation image with the SSD Inception v2 model. False positives are shown in red, false negatives are yellow, true positives are green, and blue rectangles denote ground truth for all true positive detections.}
\label{fig:fig_cowc}
\end{figure}

\begin{figure}[h]
\begin{center}
\includegraphics[width=0.95\linewidth]{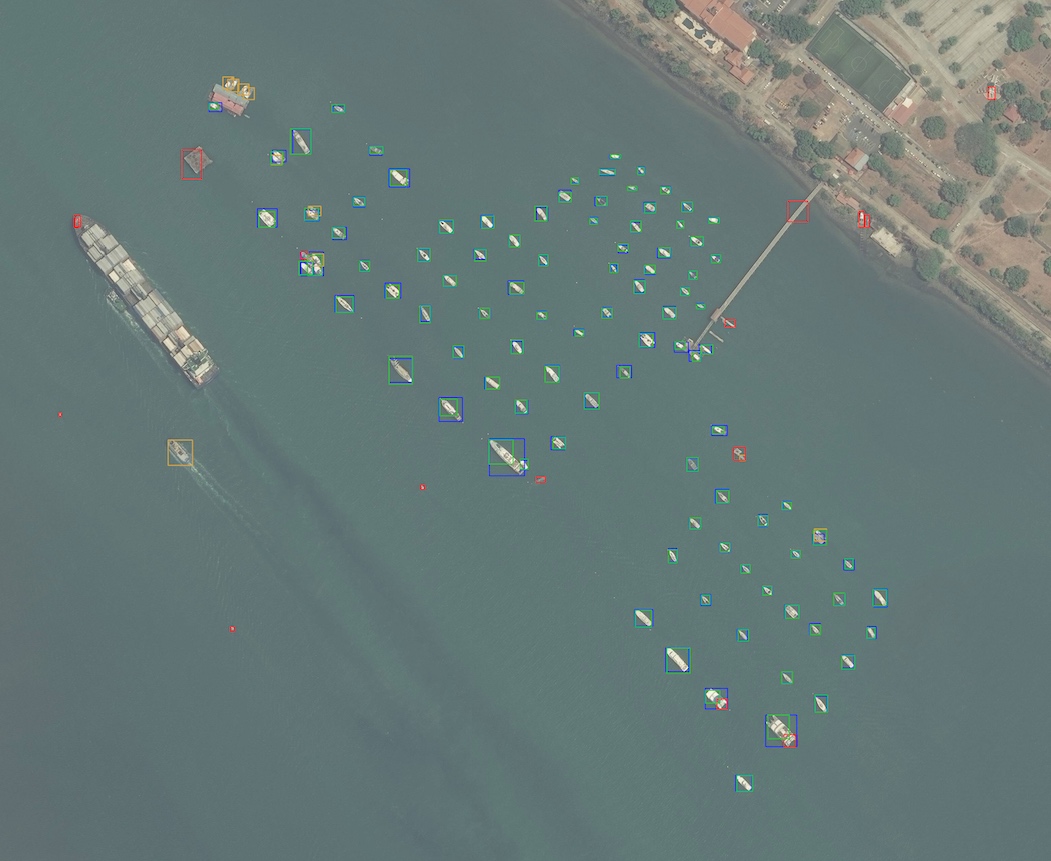} 
\end{center}
\caption{Evaluation image with the R-FCN model.  False positives are shown in red, false negatives are yellow, true positives are green, and blue rectangles denote ground truth for all true positive detections.}
\label{fig:fig_cowc}
\end{figure}

\begin{figure}[h]
\begin{center}
\includegraphics[width=0.95\linewidth]{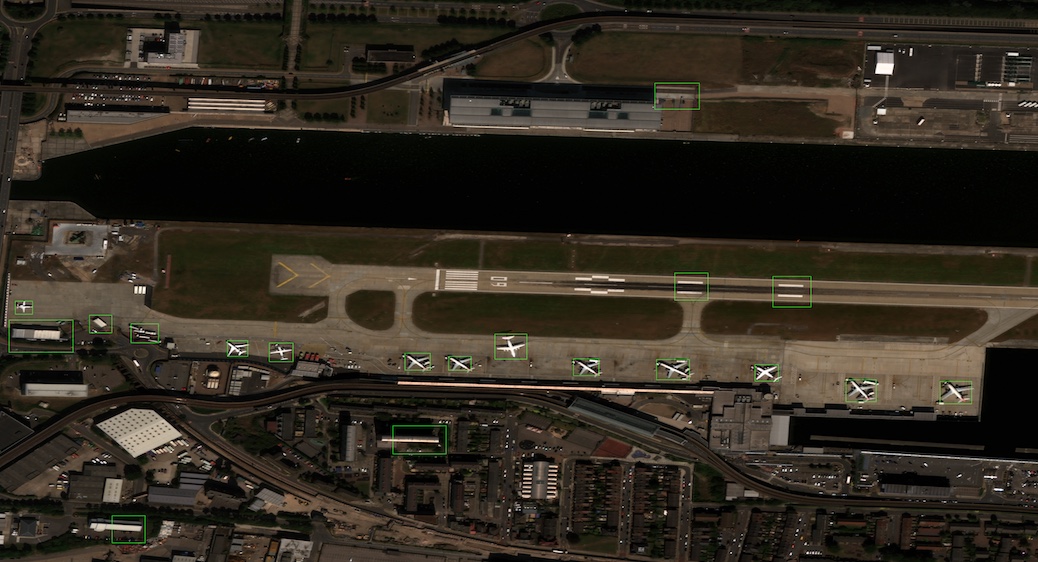} 
\end{center}
\caption{Raw detections from the Faster RCNN model at a detection threshold of 0.5.  Airplanes are shown in green; the false positive rate is high.}
\label{fig:fig_cowc}
\end{figure}

\begin{figure}[h]
\begin{center}
\includegraphics[width=0.95\linewidth]{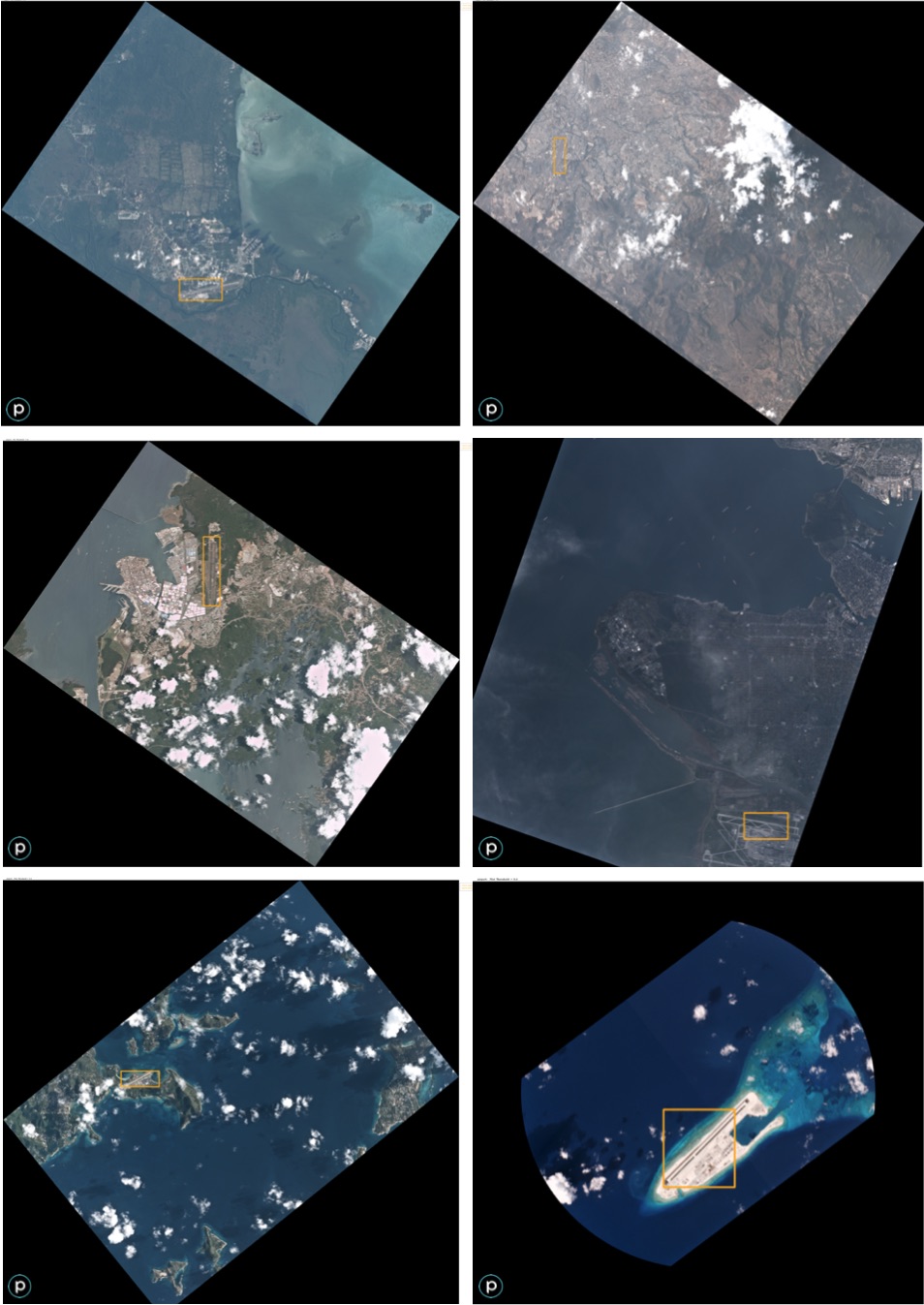}
\end{center}
\caption{Successful detections of airports and airstrips (orange) in Planet images with the YOLT model over both maritime backgrounds and complex urban backgrounds. Note that clouds are present in most images. The middle-right image demonstrates robustness to low contrast images. 
Each image takes between $1-3$ seconds to analyze, depending on size}
\label{planet_airport_success}
\end{figure}

\begin{figure}[h]
\begin{center}
\includegraphics[width=0.95\linewidth]{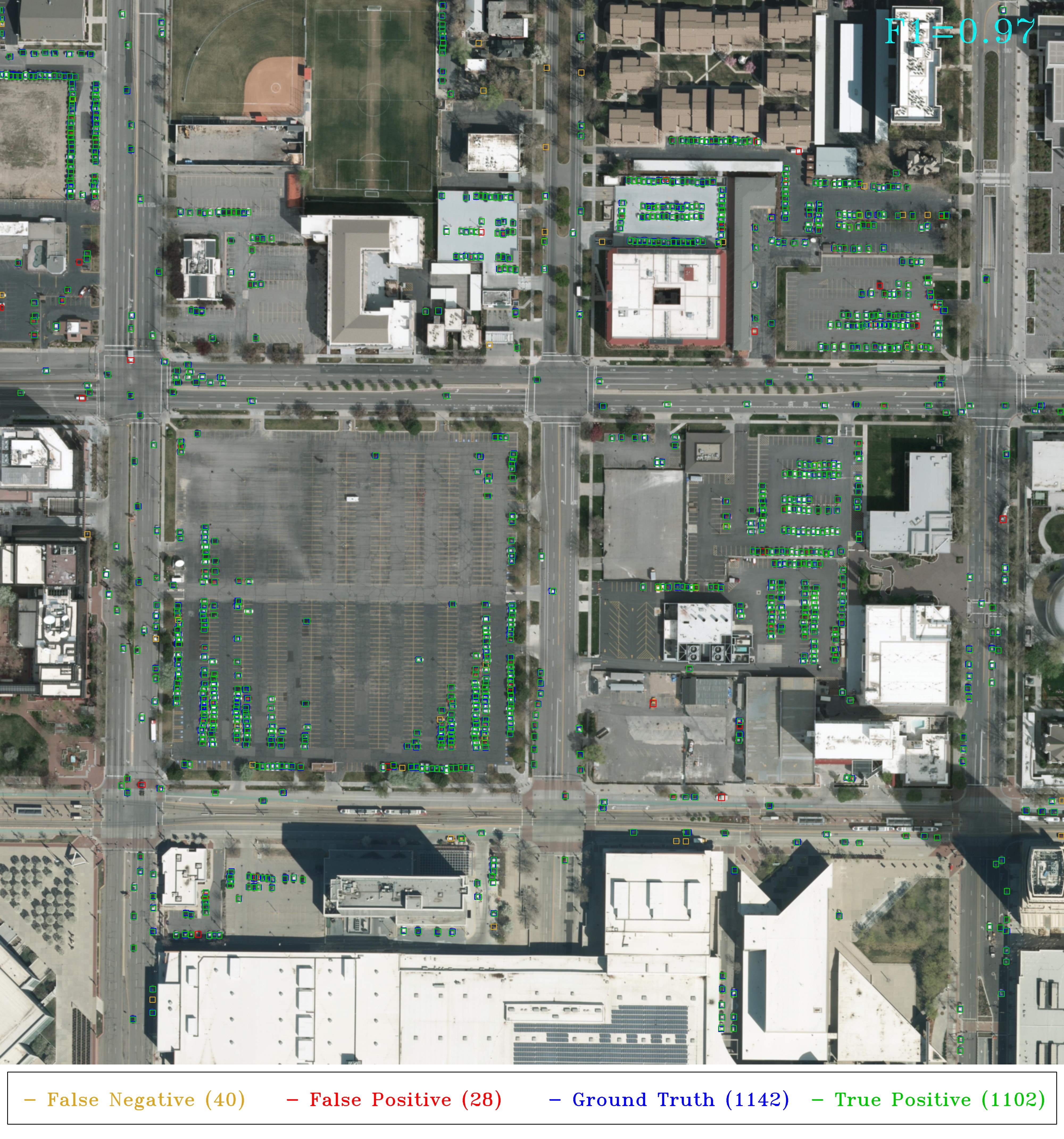} 
\end{center}
\caption{Car detection performance on a $600\times600$ m aerial image at 30 cm GSD over Salt Lake City with the YOLT model trained at 2x resolution.  F1 = 0.97 for this test image.}
\label{fig:cowc_bueno}
\end{figure}

\begin{figure}[ht]
\begin{center}
\includegraphics[width=0.95\linewidth]{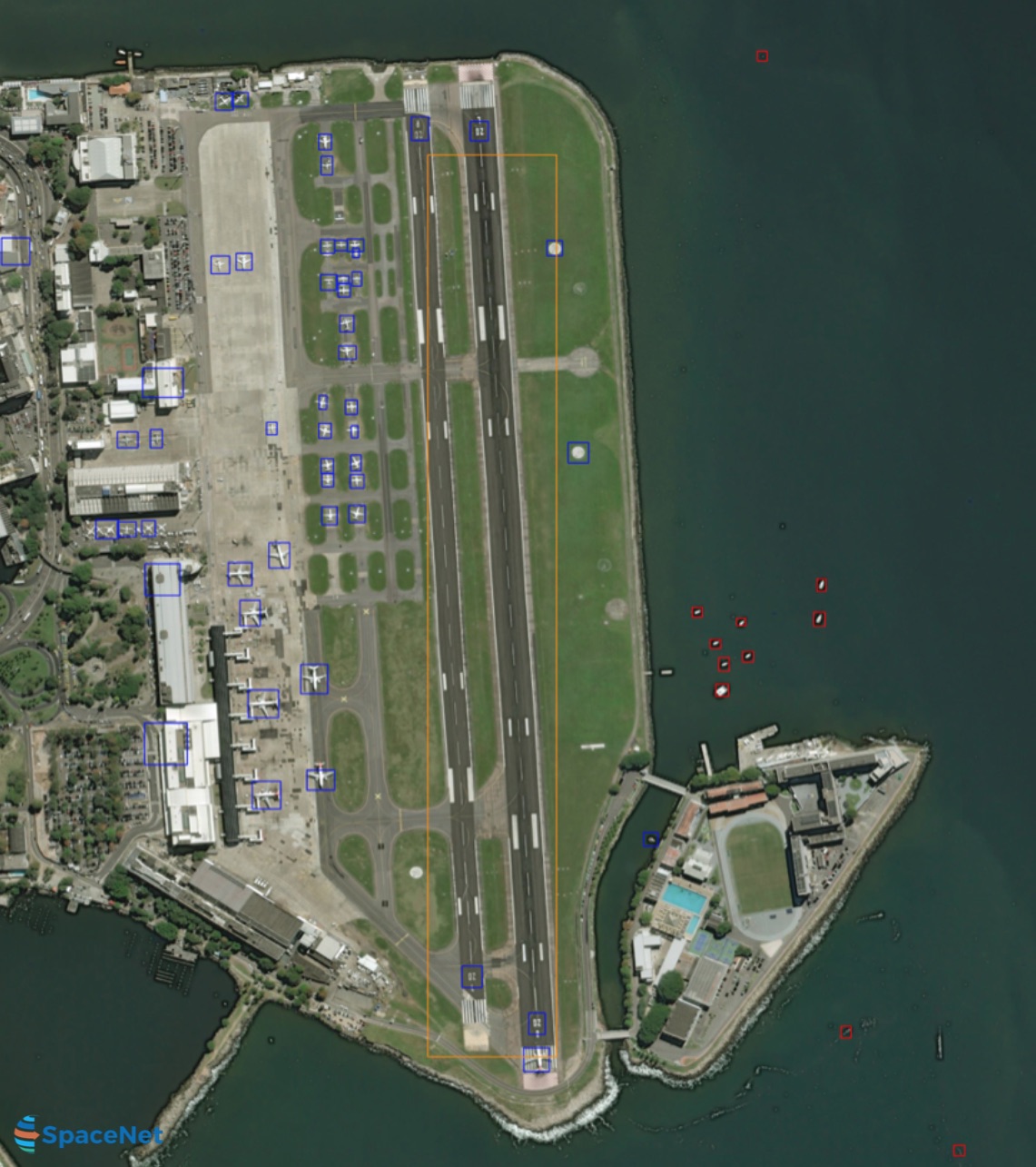}
\end{center}
\caption{SIMRDWN classifier applied to a SpaceNet DigitalGlobe 50 cm GSD image containing airplanes (blue), boats (red), and runways (orange). 
In this image we note the following F1 scores: airplanes = 0.83, boats = 0.84, airports = 1.0.}
\label{fig:fig_comb}
\end{figure}

\end{document}